\documentclass[10pt,twocolumn,letterpaper]{article}

\usepackage[pagenumbers]{cvpr} 

\usepackage{graphicx}
\graphicspath{{./images/}}
\usepackage{amsmath}
\usepackage{amssymb}
\usepackage{multirow}
\usepackage{booktabs}
\usepackage{dsfont}
\usepackage{soul, nicefrac}
\usepackage[shortlabels]{enumitem}

\usepackage[dvipsnames]{xcolor}

\usepackage{eucal}

\usepackage{caption}
\definecolor{citecolor}{HTML}{0071bc}
\usepackage[pagebackref,breaklinks=true,letterpaper=true,colorlinks,citecolor=citecolor,bookmarks=false]{hyperref}

\usepackage[capitalize]{cleveref}
\crefname{section}{Sec.}{Secs.}
\Crefname{section}{Section}{Sections}
\Crefname{table}{Table}{Tables}
\crefname{table}{Table}{Tables}

\DeclareMathOperator*{\argmax}{arg\,max}

\def\confName{CVPR}
\def\confYear{2022}

\def\prob{{\mathbf{P}}}
\def\f{{\mathbf{f}}}
\def\x{{\mathbf{x}}}
\def\z{{\mathbf{z}}}
\def\i{{\mathbf{i}}}
\def\t{{\mathbf{t}}}
\def\ImageEncoder{\mathbf{E}}
\def\TextEncoder{\mathbf{T}}
\def\I{\mathcal{I}}
\def\T{\mathcal{T}}
\def\Z{\mathcal{Z}}
\begin{document}

\title{Retrieval Augmented Classification for Long-Tail Visual Recognition\thanks{Part of this work was done when WY was with Amazon and CS was with The University of Adelaide.}  
}
\author{
Alexander Long$ ^{1,3}$, ~~ Wei Yin$ ^{2}$, ~~ Thalaiyasingam Ajanthan$ ^{1}$,
Vu Nguyen$ ^{1}$, ~~ 
Pulak Purkait$ ^{1}$, ~~ \\
Ravi Garg$ ^{1}$, ~~
Alan Blair$ ^{3}$, ~~
Chunhua Shen$ ^{4}$, ~~
Anton van den Hengel$ ^{1,2}$ 
\\[0.25cm]
\normalsize 
$ ^1$ Amazon  ~~~ $ ^2$ The University of Adelaide, Australia ~~~ 
$ ^3$ University of New South Wales ~~~
$ ^4$ Zhejiang University, China 
}

\maketitle
\begin{abstract}
   We introduce Retrieval Augmented Classification (RAC), a generic approach to augmenting standard image classification pipelines with an explicit retrieval module. RAC consists of a standard base image encoder fused with a parallel retrieval branch that queries a non-parametric external memory of pre-encoded images and associated text snippets. We apply RAC to the problem of long-tail classification and demonstrate a significant improvement over previous state-of-the-art on Places365-LT and iNaturalist-2018 ($14.5\%$ and $6.7\%$ respectively), despite using only the training datasets themselves as the external information source. We demonstrate that RAC's retrieval module, without prompting, learns a high level of accuracy on tail classes.
   This, in turn, frees the base encoder to focus on common classes, and improve its performance thereon. RAC represents an alternative approach to utilizing large, pretrained models without requiring fine-tuning, as well as a first step towards more effectively making use of external memory within common computer vision architectures.

\end{abstract}

\section{Introduction}
Large Transformer \cite{transformer} models have arrived in Computer Vision, with parameter counts and pretraining dataset size increasing rapidly \cite{vit, clip, cvtransformers, jft, resnext, scalingvits}. The distributed representations learned by such models result in significant performance gains on a range of tasks, however come with the drawback of storing world knowledge implicitly within their parameters, making post-hoc modification\cite{lmediting} and interpretability \cite{vitinterp} challenging. In addition, real-world data is long-tailed by nature, and implicitly storing every visual cue present in the world appears futile with current hardware constraints. As an alternative to this fully parametric approach, we propose augmenting standard classification pipelines with an explicit external memory, thus separating model performance from parameter count, and facilitating the dynamic addition and removal of information explicitly with no changes to model weights. 

To evaluate our approach, we focus on the problem of Long-Tail visual recognition, as it shares many of the properties likely to be encountered by a general agent. Specifically, the data distributions are highly skewed on a per-class basis, with a majority of classes containing a small number of samples.
The number of samples in these small classes, commonly referred to as the ``tail'', can far outweigh those in the relative minority of high sample classes (referred to as the ``head''). In this situation, learning is challenging due to both the lack of information provided for tail classes, and the tendency for head classes to dominate the learning process. Long-tail learning is a well-studied~\cite{lt1, lt2, lt3} instance of the more general label shift problem~\cite{datasetshift}, where the shift is static and known during both training and testing. Despite being well-studied, commonly occurring, and of great practical importance, classification performance on long-tail distributions lags significantly behind the state-of-the-art for better balanced classes~\cite{decoupling}.

Base approaches are largely variants of the same core idea---that of ``adjustment'', where the learner is encouraged to focus on the tail of the distribution.  This can be achieved  implicitly, via over/under-weighting samples during training~\cite{smote, bsmote, undersampling, huang2016learning} or cluster-based sampling~\cite{paco}, or explicitly via logit~\cite{pml, lasm, ala} or loss \cite{Hong_2021_CVPR, lasm} modification. Such approaches largely focus on \textit{consistency}, ensuring minimizing the training loss corresponds to a minimal error on the known, balanced, test distribution. 

\begin{figure*}[t]
\centering 
    \includegraphics[width=0.87\textwidth]{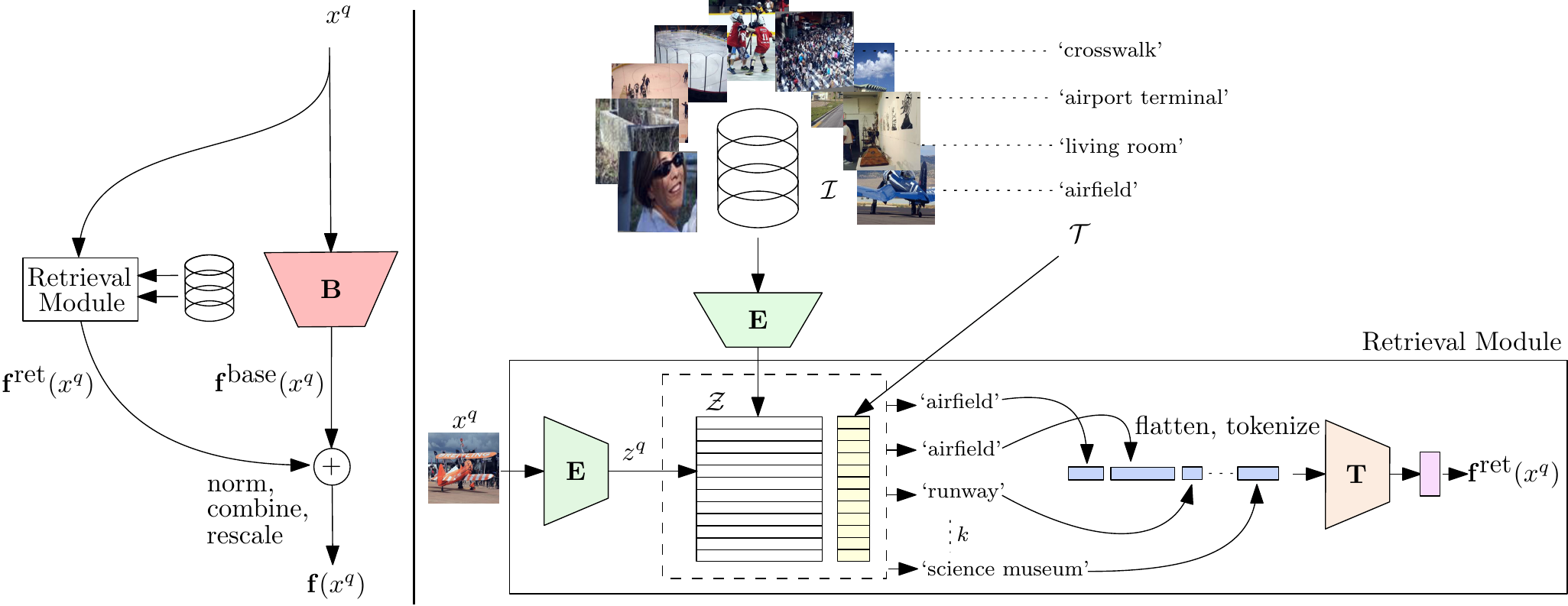} \\ 
{\small (a) The proposed RAC architecture ~~~~~~~~~~~~~~~~~~~~~~~~~~~~~~~~~~~~~~~~~~~~~~~~~~~~~~~~~~~~~~~~~~~~~~~ (b) Retrieval module ~~~~~~~~~~~~~~~~~~~~~~~~~~~~~~~~~~~~~~~~~~~~~~~~~~~~} 
    \caption{(a) RAC overview. RAC consists of a retrieval module that augments a standard encoder $\mathbf{B}(\cdot)$ with explicit external memory. (b) The retrieval module consists of external images $\mathcal{I}$ encoded by a fixed, pretrained image encoder $\ImageEncoder(\cdot)$, and associated text $\mathcal{T}$ queried using an approximate $k$-NN and encoded via a text encoder $\TextEncoder(\cdot)$. The logits of the retrieval encoder are then combined with those of the base network. In our instantiation, $\mathbf{B}$ and $\mathbf{E}$ are ViT's, and $\mathbf{T}$ is a BERT-like text encoder.}
    \label{fig:main}
\end{figure*}

An alternative approach focuses on ensembling models. Instead of disregarding knowledge of the test distribution, recent work \cite{guo2021long, bbn, xiang2020learning} use ensembling models to induce invariance to the test distribution. This is typically done by training separate models under different losses or re-sampling techniques, and combining them at test time.

We introduce a third approach, Retrieval Augmented Classification (RAC), motivated by the desire to explicitly store tail knowledge, as a retrieval-based augmentation to standard classification pipelines. 

RAC's retrieval module is multi-modal, making use of image representations as retrieval keys, and returning encoded textual information associated with each image. We place no limitation on the nature of this text; it may be the labels from a supervised training set, descriptions, captions etc. In the simplest case, the images in the index, and associated text, can be the images and labels from the dataset of interest alone. 

RAC jointly trains a standard base encoder, and a separate retrieval branch. We demonstrate empirically that the retrieval branch learns, without explicit prompting, to focus on tail classes.  This frees the base encoder from modelling these sparse classes, as they are already effectively represented by the non-parametric memory of the retrieval module. This in turn allows the base encoder to achieve a higher level of performance on the head classes. 

RAC achieves state-of-the-art performance on common Long-Tail classification benchmarks, even out-performing approaches such as LACE \cite{lasm} that are provably consistent with regard to the class-balanced error, and Bayes-optimal under Gaussian class priors.
\textit{ A major benefit of RAC is its ability to use large, pretrained models for inference} (for index and retrieval encoding), leveraging their rich representations to improve the classification performance of a base learner. This broadens the applicability of such models due to the large cost of fine-tuning. 

Our contributions are summarised as follows:
\begin{enumerate}
\itemsep -0.1cm
    \item The first demonstration of effective external memory within long-tail visual recognition setting.
    \item A novel method for Long-Tail Classification that significantly improves on the current state-of-the-art.
    \item Insight into the proposed method, with the reimplementation of strong baselines that also exceed current state-of-the-art. 
\end{enumerate}

\section{Related Work}
\noindent
\textbf{Resampling and Logit Adjustment}
Over-sampling sparse classes \cite{smote, bsmote} is one of the oldest approaches to addressing distribution bias, but one that is still in common use. Under-sampling common classes~\cite{undersampling}, applying additional data-augmentation to sparse classes in pixel, or feature space~\cite{featureaug, featureaug2}, or sampling uniformly from pre-computed clusters~\cite{huang2016learning}, have also been suggested. 
Hong \etal~\cite{Hong_2021_CVPR} propose a distribution aware weight regularizer that is applied more heavily to head classes than tail classes, in a similar vein to weight normalization. However, empirically, the resulting model (LADE) only produces marginal gains over straight-forward balanced softmax. 
Recent work \cite{lasm} has unified many empirically successfully approaches under a Fisher-consistent scorer for the balanced error, and additionally shown that weight normalization fails when used with the ADAM optimizer. 
Zhang \etal~\cite{DisAlign} adopt a two stage approach, and propose a class-specific learnable (from the samples) reweighting (via a single layer NN) of the frozen pretrained logits based on a generalized formulation of the class-balanced softmax. They show that transforming the classification head, as opposed to re-training it, performs better. PaCo \cite{paco}, the current state-of-the-art for long-tail  classification, combines learnable logit adjustment with contrastive learning\cite{infonce}. Despite their simplicity, adjusted logit methods (LACE, LDAM, LADE) remain strong solutions to the long tail problem, typically achieving within $1$-$2$\% top-1 accuracy of state-of-the-art ensemble approaches (see \cref{tbl:inat}, \cref{tbl:places}).   

\noindent
\textbf{Ensemble Methods}
In proposing TADE~\cite{tade}, Zhang \etal explicitly train three heads with standard, balanced, and inversely weighted softmax losses, linearly combining their predictions at test time, weighted by a measure of confidence derived from each head's stability under data-augmentation. Wang \etal ~\cite{ride} in contrast combine multiple independently trained classification heads that are pushed to be decorrelated in their predictions via a (class balanced) KL loss, with a small routing network that improves computational efficiency during inference. 

\noindent
\textbf{External Memory}
One of the first models to successfully combine deep networks with external memory was the Neural Turing Machine~\cite{NTM}.  The purpose of that model was symbolic manipulation, however, which renders its architecture quite different to that of RAC.  
Gong \etal~\cite{gong2019memorizing} proposed a similar retrieval-module architecture for anomaly detection, but without RAC's corresponding base module.
Recently, in the NLP domain, several works have proposed the augmentation of large language models with a non-parametric memory to allow explicit access to external data \cite{rag, realm}. While such approaches make use of differential retrievers, which introduces the problem of lookup/representation drift, they are still closely related to RAC. $k$-NN Language Models (LMs) \cite{knnLMs} are most similar to our work, which directly interpolate a retrieval distribution with the next token distribution produced by a base LM, resulting in reduced combined model perplexity.

Latent retrieval has been applied to textual open-domain QA\cite{latentret, dpr}. The central difference is such approaches return information that is most similar to the retrieval key, whereas RAC returns information (text) attached to retrieved samples. An approach similar to that of RAC has been applied to knowledge-intensive QA \cite{NLPSymbol}, where a `fact memory' consisting of triples from a symbolic Knowledge Base (KB) is directly encoded and queried using the final representation of a language model as keys.  In computer vision, non-parametric retrieval has been used to assist in addressing the fine-grained retrieval problem, such as in enforcing instance-level retrieval loss in \cite{grafit}. The Open-world Long-tail model proposed in~\cite{owlt} also makes use of a retrieval module, but primarily as a mechanism to distinguish between seen, and unseen samples in the `open world' setting, not to boost performance on seen classes as we do.

\section{Method}

\subsection{Preliminaries} 
In long-tailed visual recognition, the model has access to a set of $N$ training samples $\mathcal{S} = \{(\x_n, y_n)\}^N_{n=1}$, where $\x_n \in \mathcal{X} \subset \mathbb{R}^D$ and labels $\mathcal{Y}=\{1,2,..,L\}$. Training class frequencies are defined as $N_y = \sum_{(x_n, y_n) \in\mathcal{S}} \mathds{1}_{y_n=y}$  and the test-class distribution is assumed to be sampled from a uniform distribution over $\mathcal{Y}$\footnote{While this is true for Places365-LT, iNaturalist2018 has a fixed number of test samples for each class ($N^{\text{test}}_i=3, \quad \forall i\in\mathcal{Y}$)}, but is not explicitly provided during training. The goal is thus to minimize the balanced error, of a scorer $\f:\mathcal{X}\rightarrow\mathbb{R}^L$, defined as;
\begin{equation}
    {\rm BE}(\x,\f(\cdot)) = \sum_{y\in\mathcal{Y}}  \prob_{\x|y} \left(y \notin \argmax_{y' \in \mathcal{Y}} \f_{y'}(\x)  \right)
\end{equation}
where $\f_y(x)$ is the logit produced for true label $y$ for sample $\x$. Traditionally this is done by minimizing a proxy loss, the Balanced Softmax Cross Entropy (BalCE):
\begin{equation}
    \ell_{\text{BalCE}}(\x,y, \f_y(\cdot))= -\frac{1}{N_y} \log \frac{e^{\f_{y}(\x)}}{\sum_{y^{\prime} \in\mathcal{Y}} e^{\f_{y^{\prime}}(\x)}}.
\end{equation}
This is a form of \textit{re-weighting}, where the contribution of each label's individual loss is scaled by an approximation of $\prob(y)$, which for $\ell_{\text{BalCE}}$, is the inverse class frequency. BalCE remains a strong baseline in this domain (see \cref{sec:ablation}).

\subsection{LACE Loss} 
An alternative to re-weighting is to adjust the logits themselves, however the two can be done in conjunction, resulting in the general form of the re-weighted (via $\alpha_y$) and adjusted (via $\Delta y$) softmax cross entropy loss;
\begin{equation} \label{eq:general}
\ell(\x,y, \f_y(\cdot)) = -\alpha_{y} \log \frac{e^{\f_{y}(\x)+\tau \cdot  \Delta y}}{\sum_{y^{\prime} \in\mathcal{Y}} e^{\f_{y^{\prime}}(\x)+\tau \cdot \Delta y'}}
\end{equation}
where $\tau$ is a constant temperature scaling parameter. Several recent works focusing on long-tail learning exploit special cases of this loss. If $\alpha_y = 1$ and, Logit Adjusted Cross-Entropy (LACE)~\cite{lasm}, can be recovered with $\Delta_{y }=\log \left({N_y}/N\right)$ and LDAM \cite{LDAM} with $\alpha_y = \nicefrac{1}{N_y}$ and
\begin{equation}
    \Delta y = 
    \begin{cases}
N_y^{-1/4}& \text{if} \quad y'=y,\\
0 & \text{otherwise.}
    \end{cases}
\end{equation}
Both are Fisher-consistent with respect to the balanced loss. The optimal $\tau$ can be found with a holdout set, or set to 1 if the logits are calibrated \cite{calibration}. In our experiments, this calibration is achieved through label smoothing~\cite{whendoeslshelp}, which has been shown to \textit{implicitly} calibrate neural networks\cite{labelsmoothing}. This property is important in the design of RAC as we use LACE as the base loss and do not apply manual temperature adjustment, setting $\tau=1$, unless otherwise specified.

\subsection{Retrieval Augmented Classification}
The overall idea of RAC is very simple---Split the scorer into two branches (see Fig.~\ref{fig:main}), where one branch (retrieval) exhibits implicit invariance to class frequency. The two branches are trained under a common LACE loss, with their individual logits combined with a norm, addition and re-scale operation to ensure 
that 
one does not override the other during training. 

The base branch encoder $\mathbf{B}(\cdot)$ can be any choice of a standard backbone network.
In our experiments, we primarily use the ViT-B-16 variant of Visual Image Transformer~\cite{vit}, transforming the final token embedding via a standard linear layer. The retrieval module (see Sec.~\ref{sec:ret}) takes a raw image and performs a latent-space lookup on an index of precomputed embeddings, returning the text attached to the top $k$ most similar images to the image currently being considered, $\x^q$. This text is then fed through a text encoder and transformed by another linear layer into logits $\f^{\text{ret}}(\x^q)$. 

\begin{figure}[b!]
\centering
\includegraphics[width=\columnwidth]{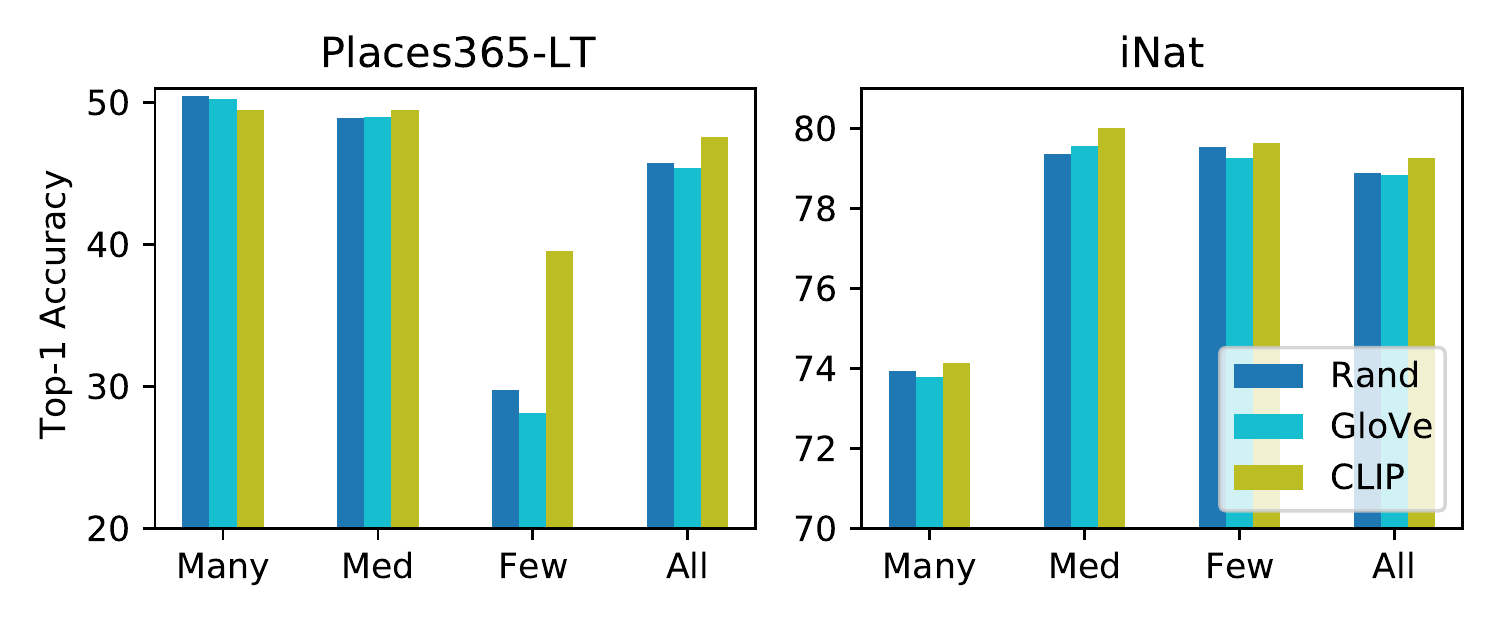}
\caption{Effect of the choice of text encoder on performance. The overall impact is minor, however the CLIP LM significantly boosts few-shot performance on Places365-LT, where labels are natural language terms.}
\label{fig:textenc}
\end{figure}

We make use of the pretrained BERT-like text encoder (63M parameters, 12-layer 512-wide model with 8 attention heads) from CLIP \cite{clip}, which we choose due to the broad (400M) range of images, alt-text pairs used during pretraining, and the compatibility with other language models due to the preservation of masked self attention in the architecture. In our experiments, the choice of text encoder is not critical as the textual information being retrieved (labels) is not highly complex, and off-the-shelf word embeddings, and even random encodings still perform reasonably well (see \cref{fig:textenc}). This choice does increase training time due to the larger parameter count (see \cref{tbl:runtime}), but allows RAC to scale to more complex retrieved text.

\begin{figure}
\centering
\includegraphics[width=\columnwidth]{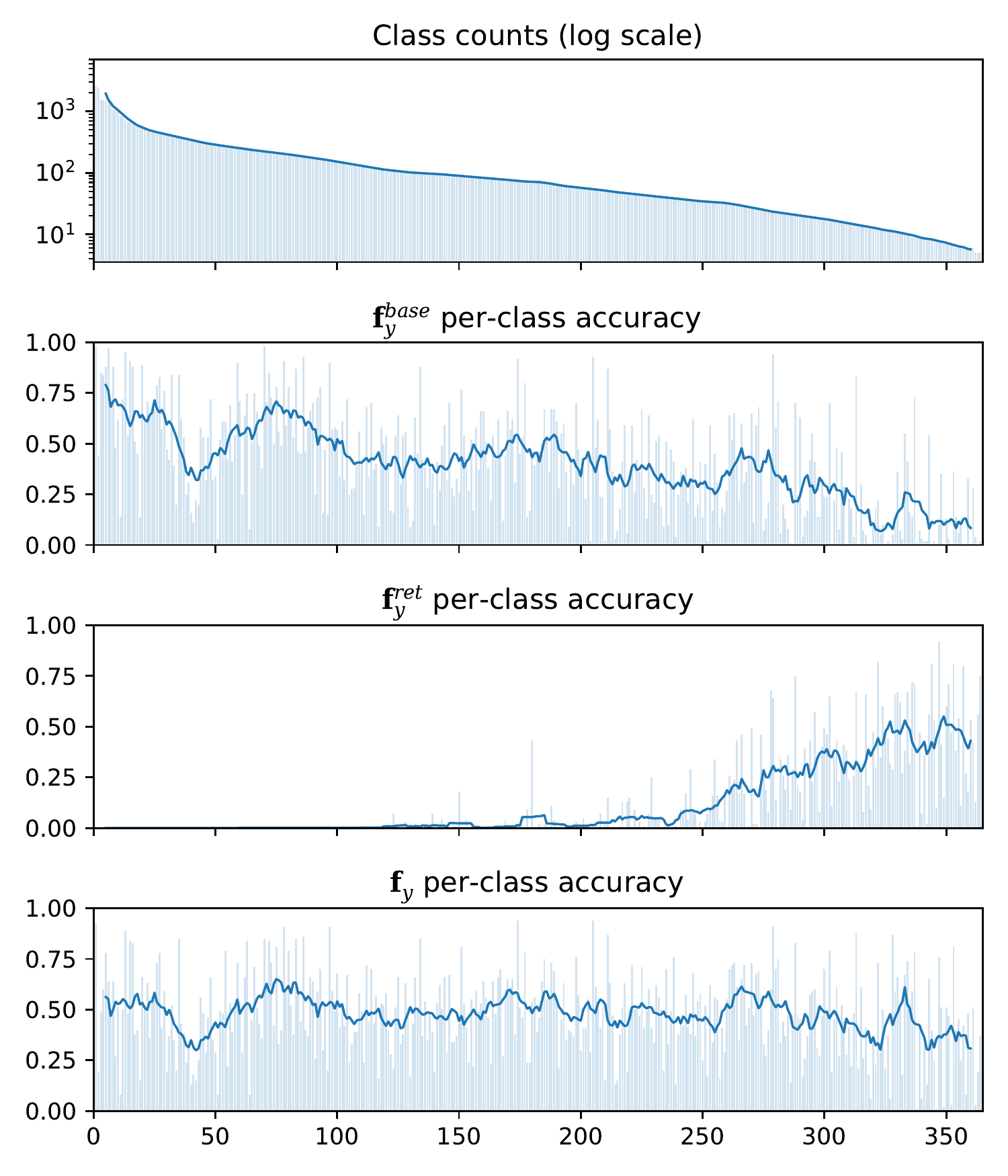}
\caption{Per-class top-1 accuracy on Places365-LT from each branch's output. Without prompting, the retrieval module learns to focus on tail classes. The $20$ sample moving average over classes (solid line) is shown for clarity.} 
\label{fig:Per-class}
\end{figure}

To combine base and retrieval branches, we normalize each branch's outputs to the unit norm and add them together. To ensure training dynamics are not altered (via lower logit magnitudes) in comparison to the baselines we rescale the combined logits by a constant factor (dependent on $L$ due to final layer Xavier initialization~\cite{glorot2010understanding} also being dependent on $L$).
\begin{equation}
    \f(\x) = \frac{L}{2} \left(  \frac{\f^{\text{ret}}(\x)}{||\f^{\text{ret}}(\x)||_2} + \frac{\f^{\text{base}}(\x)}{||\f^{\text{base}}(\x)||_2} \right),
\end{equation}
where  $\f^{\text{base}}(\x)$ represents the logits produced by the image encoder backbone, and $\f^{\text{ret}}(\x)$ is the output of the retrieval module. This straightforward setup has the benefit of being able to treat the branch outputs as individual logits, increasing the interpretability of RAC, and allowing us to precisely evaluate the per-class accuracy of each branch (see \cref{fig:Per-class}). While there are many ways to combine the branches such as confidence or distance based weightings, attention mechanisms etc., we found this approach sufficient, with the weighting of each branch done implicitly by the learned sharpness of the logits.

\subsection{Retrieval Module}\label{sec:ret}

The retrieval module consists of a frozen pretrained image encoder $\ImageEncoder(\cdot)$, a pre-existing set of external images $\I = \{\i_j\}_{j=1}^J$, with associated text $\T = \{\t_j\}_{j=1}^J$, which may be labels, descriptions, captions etc. Unless otherwise specified $\ImageEncoder(\cdot)$ is a ViT-B-16, pretrained on ImageNet following \cite{howtotrainvit}. Prior to training RAC, the retrieval module is initialized by producing image keys $\Z = \{\z_j\}$ such that $\z_j = \ImageEncoder(\i_j) \; \forall j$, and storing the resultant representations in a fast approximate $k$-NN index.

During training, we produce features $\z^q = \ImageEncoder(\x^q)$ for each image $\x^q$ in the training batch.  The $k$-NN is queried for each $\z^q$ and returns a list of indices of the $k$ closest keys in $\Z$, where cosine similarity is the distance metric. The text element in $\T$ is recovered for every such index, generating $k$ text elements for each query. These text elements are then encoded by a text encoder $\TextEncoder(\cdot)$ 
which produces the retrieval branch's (fixed length) logits, $\f^{\text{ret}}(\x^q)$.

Text strings are truncated after 76 tokens, and the resultant batches are zero-padded. This approach allows for a single text-encoder call per batch, as opposed to $k$ which would be required if each text snippet was encoded separately, and would result in a significant slowdown. The use of a large-scale transformer ensures that RAC can scale to longer text snippets if the external information is expanded to contain additional sources beyond simply labels. 

\textit{A key feature of the retrieval module is its ability to include otherwise unconnected data-sources simply via their labels.} 
In this way, we can dynamically add or remove datasets from $\mathcal{\I}$, and if new examples are similar (from the point of view of the encoder), they can directly impact classification accuracy, providing an alternative to fine-tuning in order to incorporate new information. 

For fast querying of the index, we make use of the FAISS implementation \cite{faiss} of the Hierarchical Navigable Small World (HNSW) approximate $k$-NN lookup \cite{hnsw}. We construct the index with default settings aside from the hyperparameter $M=32$, which sets the number of bidirectional links per node and increases the complexity of the index, but allows for higher recall. During training, we drop the first result, as when training data is included in the index, the first result is often the original image, which causes the text encoder to place undue weight on the first retrieved label when creating predictions.

\def\x{{$\times$}}
\begin{table}[]
\centering
\begin{tabular}{@{}lllll@{}}
\toprule
\textbf{Method}               & \textbf{Many}   & \textbf{Med}    & \textbf{Few}    & \textbf{All}    \\ \midrule
\multicolumn{5}{c}{\textbf{Input: $224\times 224$}}                                                           \\ \midrule
OLTR   \cite{owlt} $\dagger$                  & 59              & 64.1            & 64.9            & 63.9            \\
Decouple-LWS   \cite{decoupling} $\dagger$    &  -               &   -              &     -            & 65.9            \\
LADE   \cite{Hong_2021_CVPR} $\dagger$                  & -                &    -             &   -              & 69.3            \\
Grafit \cite{grafit}                       &   -              &   -              &  -               & 69.9           \\
ALA \cite{ala}                       & 71.3            & 70.8            & 70.4            & 70.7            \\
RIDE   \cite{ride} (2 experts)       & 70.2            & 71.3            & 71.7            & 71.4            \\
LACE   \cite{lasm}                   &  -               &  -               &  -               & 71.9           \\
RIDE   \cite{ride} (4 experts)       & 70.9            & 72.4            & 73.1            & 72.6            \\
TADE   \cite{tade}                   &  74.4	        & 72.5	        & 73.1              & 72.9            \\
DisAlign \cite{DisAlign}       & - & -& - & 74.1 \\
PaCo \cite{paco}                  &   75.0	            & 75.5	        & 74.7	            & 75.2           \\
RAC  (ours)                         & \textbf{75.92}  & \textbf{80.47} & \textbf{81.07} & \textbf{80.24} \\ \midrule
\multicolumn{5}{c}{\textbf{Input: $384\times 384$}}                                                            \\ \midrule
Grafit                        &   -              & -                &  -               & 81.2            \\
RAC  (ours)                         & \textbf{82.91} & \textbf{85.71} & \textbf{86.06} & \textbf{85.56} \\ \bottomrule
\end{tabular}
\caption{Historical performance on iNat, $^\dagger$Results reproduced from \cite{ala}.}
\label{tbl:inat}
\end{table}

\begin{table}[]
\centering
\begin{tabular}{@{}lllll@{}}
\toprule
\textbf{Method}           & \textbf{Many}  & \textbf{Med}   & \textbf{Few}   & \textbf{All}   \\ \midrule
Focal Loss \cite{focal} $\dagger$      & 41.1           & 34.8           & 22.4           & 34.6           \\
Range Loss  \cite{range} $\dagger$      & 41.1           & 35.4           & 23.2           & 35.1           \\
OLTR   \cite{owlt} $\dagger$             & 44.7           & 37             & 25.3           & 35.9           \\
Decouple-LWS   \cite{decoupling} $\dagger$ & 40.6           & 39.1           & 28.6           & 37.6           \\
LADE   \cite{Hong_2021_CVPR} $\dagger$             & 42.8           & 39             & 31.2           & 38.8           \\
DisAlign \cite{DisAlign}  & 40.4  & 42.4  & 30.1  & 39.3 \\
ALA \cite{ala}                   & 43.9           & 40.1           & 32.9           & 40.1           \\
TADE \cite{tade}     &     43.1 &	42.4 &	33.2 & 40.9 \\

PaCo \cite{paco}                 & 36.1           & 47.9           & 35.3           & 41.2           \\ \midrule
RAC (ours)              & \textbf{48.69} & \textbf{48.31} & \textbf{41.76} & \textbf{47.17} \\ \bottomrule
\end{tabular}
\caption{Historical performance on Places365-LT. $^\dagger$Results reproduced from \cite{ala}.}
\label{tbl:places}
\end{table}

\section{Experiments} \label{sec:experiments}

\begin{table}[]
\centering
\begin{tabular}{@{}llrrrr@{}}
\toprule
\textbf{Method} & \textbf{$\mathbf{B}$} & \textbf{Many} & \textbf{Med}   & \textbf{Few}   & \textbf{All} \\ \midrule
\multicolumn{6}{c}{\textbf{Places365-LT}}                                                                          \\ \midrule
CE                   & RN50     &            -    &     -           &           -     & 32.14          \\
BalCE                & RN50     &           -     &       -         &            -    & 38.31          \\
CE                   & ViT-B-16 & \textbf{50.81} & 33.83          & 19.51          & 37.16          \\
BalCE                & ViT-B-16  & 49.03         & 45.72         & 29.05         & 43.67         \\ \midrule
Retrieval & -        & 43.50           & 41.99          & 26.83          & 39.58        \\
Base      & ViT-B-16 & 44.57          & 45.06          & 40.77          & 44.05         \\
RAC                  & ViT-B-16 & 48.69          & \textbf{48.31}          & \textbf{41.76}          & \textbf{47.17}  \\ \midrule
\multicolumn{6}{c}{\textbf{iNaturalist 2018}}                                                                      \\ \midrule
CE                   & RN50            &        -        & -               &        -        & 61.7           \\
BalCE                & RN50            &         -       &      -          &        -        & 69.8           \\
CE                   & ViT-B-16     & \textbf{81.53} & 76.62          & 69.82          & 74.44          \\
BalCE                & ViT-B-16          & 72.39          & 76.06          & 73.05          & 74.49          \\ \midrule
Retrieval & -               & 50.10          & 52.77          & 52.45          & 52.37          \\
Base      & ViT-B-16          & 74.41          & 78.95          & 78.55          & 78.32          \\
RAC                  & ViT-B-16        & 75.92          & \textbf{80.48} & \textbf{81.07} & \textbf{80.24} \\ \bottomrule
\end{tabular}
\caption{Comparison of top-1 accuracy against baselines under a common training scheme. Column $\mathbf{B}$ indicates the architecture of the base branch. }
\label{tbl:ablation}
\end{table}

We establish RAC's high level of performance on common benchmark datasets iNaturalist2018 (\cref{tbl:inat}) and Places365-LT (\cref{tbl:places})\footnote{We do not compare against CIFARLT and ImageNetLT, as in these scenarios training is typically performed from scratch, and RAC requires a pretrained network for the retrieval module. While it is possible to train the base network from scratch, this is not a fair comparison and RAC significantly outperforms other methods due to the information present in $\mathbf{E}$.} with no additional external information aside from the datasets used to pretrain the individual encoders. Note that these tables report results from the literature which were obtained under varying architectures and training schemes. We ablate the benefit of RAC's improved training pipeline in \cref{tbl:ablation} where we reimplement class-balanced softmax Cross Entropy (BalCE) and LACE~\cite{lasm} as baselines. We consider `Base' trained under the LACE loss~\cite{lasm} as our primary baseline, due to LACE's strong theoretical grounding, provable consistency, and high level of previously reported empirical performance. We report overall accuracy as well as per-class accuracy bucketed into the few ($<20$), medium ($\leq100$) and many ($>100$) shot categories. The full per-class distribution curve is also shown in \cref{fig:Per-class}. We then focus specifically on the design choices of the retrieval module and how the choice of data for the index affects RAC in \cref{sec:indexcontent}. 

In all experiments, unless otherwise indicated, $\mathbf{E}$ is a ViT-B-16 encoder, with the weights from \cite{vit}. The weights are obtained from pretraining on ImageNet21k (IM21k), a larger (11M samples) variant of the original 1.2M images ImageNet \cite{imagenet} dataset, with more granular classes. We make use of IM21k to expand the index in some experiments, and use the variant introduced in \cite{bigtransfer}.

\subsection{Places365-LT}
Places365-LT is a synthetic long-tail variant of Places-2 \cite{places} introduced in \cite{owlt}. It consists of $365$ high-level scene classes such as `airport', `basement', etc.\
across $62.5$K samples at $256 \times 256$ resolution. The minimum number of samples per class is $5$, with a training set that, while balanced, is not perfectly uniform. The dataset contains a significant amount of label noise, which makes it appealing, as logit adjustment methods typically assume fully separable classes in their theoretical motivation.

We observe that with no explicit prompting, the retrieval network learns to highly skew its accuracy towards the few-shot classes (Fig.~\ref{fig:main}), confirming our hypothesis that it will be beneficial in this case. Note that there is no explicit signal pushing the retrieval network to learn infrequent classes over common ones, or for the supervised network to prefer common classes, as both are trained under the common LACE loss. Interestingly, RAC's learned strategy is similar to the hard-coded ensembling utilized in TADE~\cite{tade}, which is the previous state-of-the-art.

\subsection{iNaturalist-2018}
iNaturalist-2018 (iNat) \cite{inat} consists of $437$K images and $6$ levels of label granularity (kingdom, genus etc.). Following other work, we consider only the most granular labels (species), which constitutes $8142$ unique classes with a naturally occurring class imbalance. In many cases the labels are very fine-grained, making it a challenging dataset even without the long-tailed property. The test set is perfectly balanced, with $3$ samples per class. 

In addition to the $224\times 224$ resolution commonly studied, we report the results with $384\times 384$ images, which was used in GRAFIT \cite{grafit} and is currently state-of-the-art for this task. We found the use of $16 \times 16$ patch size to be of major importance on iNat, boosting retrieval accuracy by $21.6\%$ (see \cref{tbl:retrieval}), likely due to the fine-grained nature of the dataset. 

\subsection{Ablation}\label{sec:ablation}

\begin{table}[]
\centering
\begin{tabular}{@{}lrrrrr@{}}
\toprule
\textbf{Encoder} & \textbf{Many} & \textbf{Med} & \textbf{Few} & \textbf{All} & 
\textbf{ CT(m:s)}
 
 \\ \midrule
\multicolumn{6}{c}{\textbf{Places365-LT}}              \\ \midrule
RN50           & 31.73 & 16.28 & 8.65  & 20.34 & 0:46  \\
RN152d         & 33.52 & 17.71 & 10.03 & 21.89 & 2:07  \\
ViT-B-32       & 38.34 & 24.82 & 15.83 & 27.92 & 0:20  \\
ViT-B-32$^*$ & 39.95 & 26.12 & 16.87 & 29.28 & 0:49  \\
ViT-B-16       & 39.97 & 26.74 & 18.65 & 29.91 & 0:53  \\
ViT-B-16$^*$ & 40.79 & 27.23 & 19.25 & 30.55 & 3:15  \\ \midrule
\multicolumn{6}{c}{\textbf{iNaturalist 2018}}          \\ \midrule
RN50           & 26.8  & 20.8  & 21.15 & 21.56 & 5:14  \\
RN152d         & 38.95 & 29.56 & 28.45 & 30.09 & 17:50 \\
ViT-B-32       & 48.14 & 43.69 & 44.1  & 44.31 & 2:35  \\
ViT-B-16       & 59.38 & 53.92 & 52.42 & 53.89 & 4:19  \\
ViT-B-16$^*$ & 66.15 & 61.54 & 60.92 & 61.77 & 22:14 \\ \bottomrule
\end{tabular}
\caption{Analysis$\,$of$\,$standard$\,$retrieval$\,$performance$\,$and$\,$Construction Time (CT) which includes both image encoding and HNSW indexing. $^*384 \times 384$ resolution variants.}
\label{tbl:retrieval}
\end{table}

We baseline RAC's performance against class-balanced softmax cross entropy (BalCE) and with the retrieval branch removed (Base only variant) under a common training setup in \cref{tbl:ablation}. We also include final accuracies for ResNet models for comparison. RAC consistently increases all-class top-1 accuracy by $8.04\%$ on Places365-LT and $7.72\%$ on iNat over BalCE, and by $25.48\%$ over standard cross entropy on Places365-LT. These improvements are most pronounced on the tail classes where RAC improves over BalCE by $30.42\%$ on Places365-LT and $10.98\%$ on iNat. 

One question is how RAC is able to so outperform methods that are provably consistent, such as LACE \cite{lasm}. We hypothesize that, in addition to the non-convexity introduced by using Neural Networks as the scorer, this is due to the fact that sample frequency alone does not indicate classification `difficulty' from the perspective of a balanced learner \cite{adaptivelogit}. Instead, a small number of samples may still define a sufficiently clear decision boundary if the volume of semantic space covered by that class is small and distinct \cite{effectivesamples}, and hence in a truly balanced model, both inter- and {intra}-class distributions must be considered. 
Accounting for
the {intra}-class distribution 
being 
difficult, however, given no prior on this quantity is typically provided, aside from the labels themselves. Instead, the majority of prior work has either ignored this factor, or assumed the class distributions to be Gaussian. In our formulation, these ``easy" classes get picked up by the retrieval model, leaving the base branch to focus on examples that are difficult, where the difficulty is a combination of presentation frequency and class complexity. Previous methods have attempted this by correlating stability under augmentation with confidence~\cite{tade}, however, this correlation is weak.

\begin{figure}
\centering
\includegraphics[width=0.98\columnwidth]{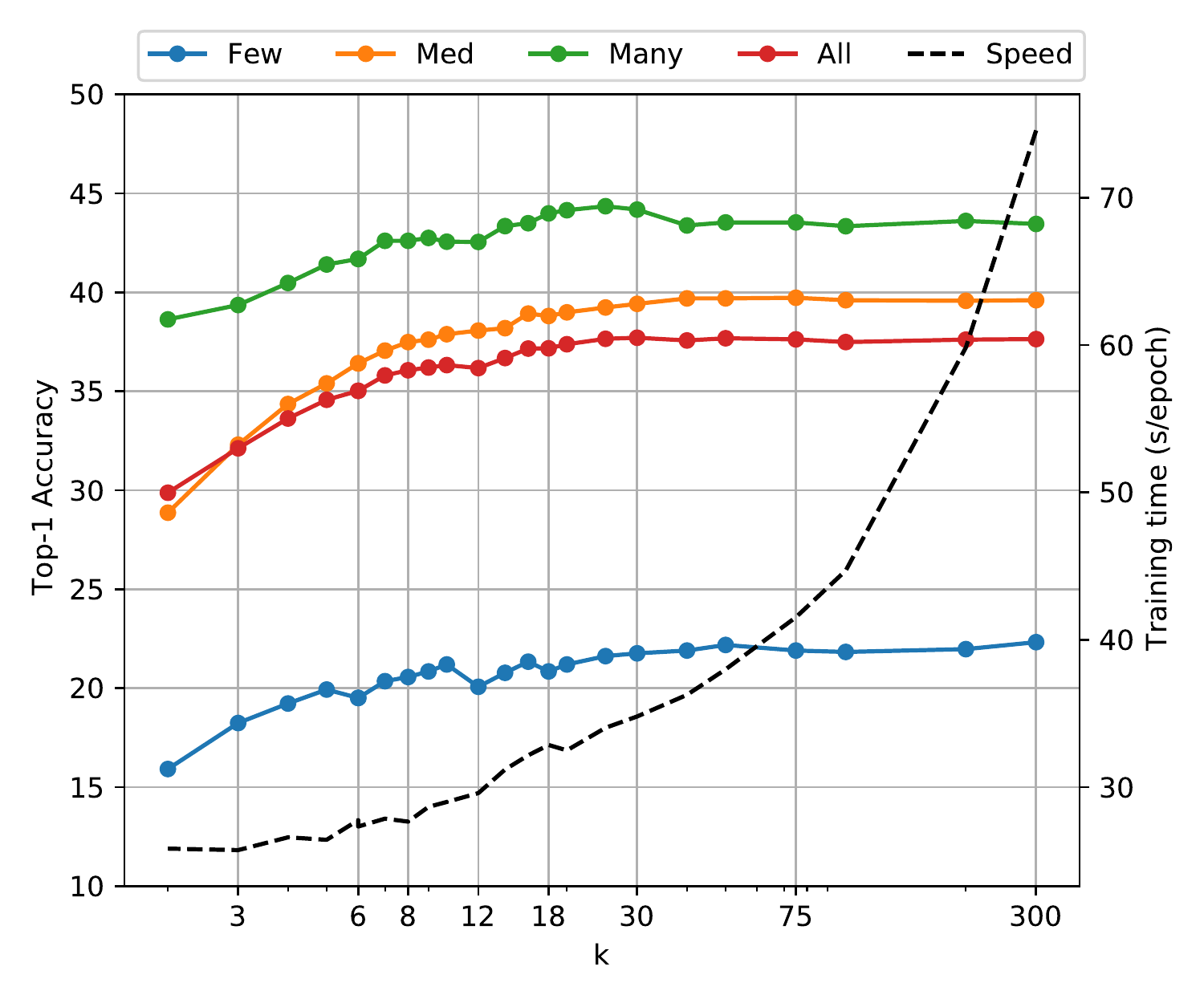}
\caption{Effect of the number of retrieved text snippets, $k$, on Places365-LT top-1 accuracy for the retrieval only branch, querying an index containing only the Places365-LT training set. Higher $k$ consistently improves performance until the cutoff induced by the text encoding truncation ($76$ tokens), however it does come at the cost of (linearly) higher training time. We choose $k=30$ in our experiments. $x$-axis is log-scaled. } 
\label{fig:k}
\end{figure}

\subsection{Retrieval}
Quantifying retrieval accuracy is important because if standard retrieval performance is significantly lower  than that of a balanced supervised learner such as the LACE baseline, it is unlikely to be beneficial. In \cref{tbl:retrieval} we perform standard retrieval with ImageNet pretrained encoders on both datasets, encoding the training set and then querying it with encoded test images, returning the label of the closest image in the training set as the prediction. All comparisons are done on exact match indexes with the $\ell_2$ distance, no data augmentation and consistent crop, interpolation and normalization constants. $z_q$ has length $2048$ for the ResNet models, and $768$ for ViTs. Despite being trained on the same data, we show that ViTs significantly outperform ResNets, and are hence critical to RAC's performance.

\subsection{Importance of the Text Encoder}

RAC makes use of a large BERT-like text encoder to learn a mapping from retrieved labels to class logits. Here we quantify the importance of this model relative to two alternatives: (i) Bag-of-words (BoW) GLoVe\cite{glove} embeddings, and (ii) BoW cached random embeddings.

Both are of dimension $300$ vs.\  $512$ for the CLIP encoder. The random embeddings are sampled from a uniform distribution over the interval $[0,1)$ and cached for each word in the input string. That is, the embeddings for individual words are consistent, but have no inherent semantics. 

We observe that a higher capacity $\TextEncoder$ does improve performance, particularly on the Places365-LT few-shot classes, but that overall this benefit is minor. This is likely due to the input to the encoder not being overly complex, and more detailed information such as captions were returned, this effect may be more pronounced.

\begin{figure*}
\centering
\includegraphics[width=0.98\textwidth]{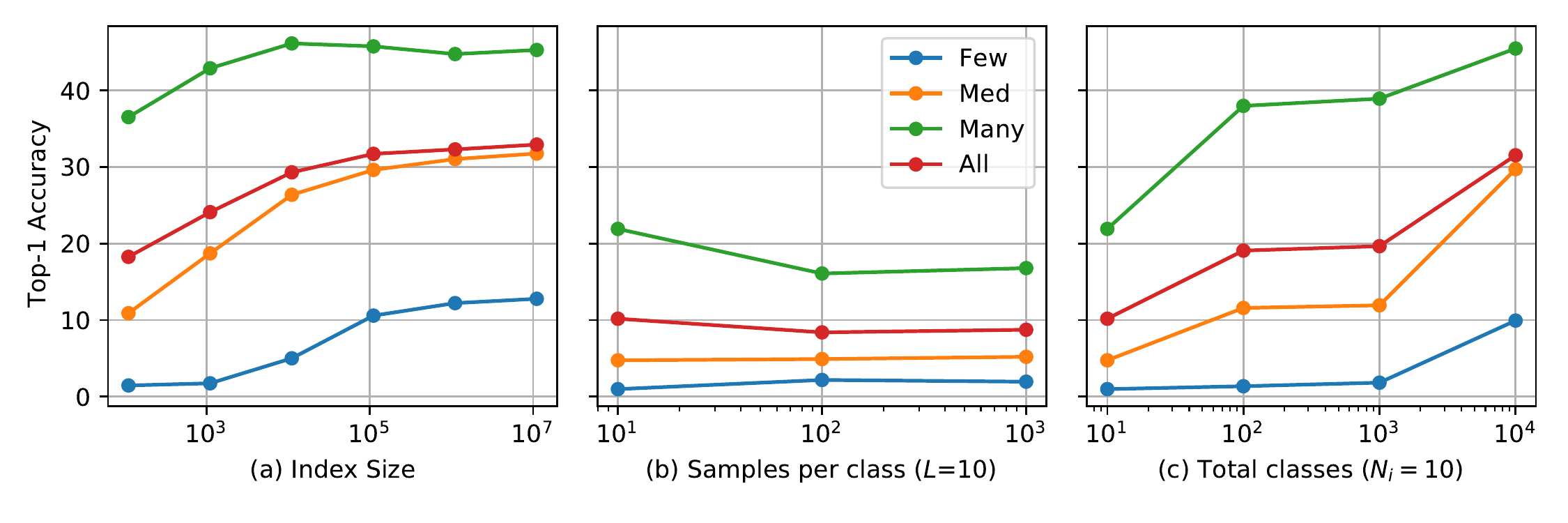}
\caption{Effect of index content on performance ($k=30$) on the retrieval branch only, trained under the LACE loss on Places365-LT. Here, the index contains no Places365-LT data, only variants of the ImageNet21k dataset. }
\label{fig:content}
\end{figure*}

\subsection{Effect of $k$}
Given that our choice of $k$ in approximate $k$-NN search is larger than the minimum number of samples present per-class for both Places365-LT and iNat, we question whether the additional returned samples, which cannot be the correct class (in the few-shot case), degrade retrieval performance. 
To study this, we experimented on only the retrieval branch, with no base encoder, and utilized an index that contained the training set only. As can be seen in Fig \ref{fig:k}, increasing $k$ consistently increases accuracy, indicating the text encoder $\TextEncoder(\cdot)$ is able to learn to disregard the common classes. Note the few-shot performance is low here, as the retrieval branch is still trained under the LACE loss, and hence pushed towards balanced performance across all classes. It is thus not free (via the base branch) to focus on the tail classes. This indicates that newer transformer architectures, that facilitate longer sequence length, may be beneficial when applied to RAC, especially when the associated text snippets themselves are longer.

\subsection{Impact of Index Content}\label{sec:indexcontent}
To quantify how index content affects RAC,  we carried out three experiments in which we trained only the retrieval module on the Places365-LT dataset, with variants of the ImageNet21k dataset used for the index. Training was done with the same final LACE loss as complete RAC, with a ViT-B-16 as $\mathbf{E}$. Specifically, we alter:
\begin{enumerate}[(a)]
\itemsep -0.1cm
    \item the index size via directly sub-sampling from the full ImageNet21k dataset.
    \item the number of training examples per-class while keeping the number of classes constant.
    \item the number of classes while keeping the number of sample per class constant.
\end{enumerate}

Results are shown in \cref{fig:content}. While naively increasing index size does increase performance, this effect diminishes as more samples are added. This is likely caused by the information content of the labels passed to $\mathbf{T}$ not increasing---as once most labels are present $\mathbf{E}$ is more likely to find a similar image, however from the perspective of $\mathbf{T}$, which is not distance or image aware, the information is the same. This is supported by sub-figures (b) and (c), in which the total amount of samples in the index is increased consistently between both plots, but adding samples by via new labels has a disproportionately larger effect than adding new samples with the number of labels constant. This is promising in that it indicates increasing label granularity, through the use of image captions or associated text,   is likely to increase RAC's performance even further.

\subsection{Runtime Consideration}
Nearest neighbour searches can be computationally infeasible for large datasets. We show here, however, that a lookup over a sample index of size $>$10M can be performed for each training sample with negligible overhead, although the additional label encoding (and subsequent backprop), does increase the training time by a factor of  $1.5-2\times$. We report the precise run-times of models with and without retrieval augmentation on Places365-LT in \cref{tbl:runtime}. Given that the index is static, the number of iterations per second is constant throughout training. All training is carried out on a single node, containing $8\times$  A$100$ GPUs ($32$GB Mem). 

Moving a tensor from the GPU to CPU, querying the index, then moving the resultant tensor back to GPU maybe expected to slowdown training. However, we find that the impact is minor with the majority of overhead coming from the additional text encoder (the random encoder, `Rand.', contains no additional parameters). To facilitate multi-node training, RAC keeps separate, complete copies of each index in memory for each node, ensuring querying the index is never the bottleneck, which we found to be essential. While it is possible to do the search entirely on GPU, due to the low overhead we do not  do this, instead using the free GPU memory to facilitate a large batch size. Due to the use of HNSW, index query time is logarithmic with respect to
the 
index size, and a standard exhaustive search is prohibitively slow.

\begin{table}[]
\centering
\begin{tabular}{@{}lrlr@{}}
\toprule
\textbf{Index Data} & \multicolumn{1}{r}{\textbf{Size}} & \textbf{Text Enc.} & \multicolumn{1}{l}{\textbf{Speed (s/epoch)}} \\ \midrule
None                    & None & None & $23.6$ \\
Places               & $184$K & Rand. & $28.3$ \\
Places               & $184$K & CLIP & $44.3$ \\
Places, IM21k & $11.2$M & CLIP & $47.0$ \\ \bottomrule
\end{tabular} \caption{Effect of additional text encoder, and lookup on training wall-time for RAC ($k=50$) on Places365-LT. Top row indicates the use of the base encoder only. The majority of added overhead comes from use of the text encoder, rather than the lookup itself.}
\label{tbl:runtime}
\end{table}

\section{Limitations}
While RAC demonstrates robust performance for both naturally occurring (iNat) and constructed (Places365-LT) long-tailed class distributions, the analysis could be further expanded to include additional long-tailed datasets. The performance of RAC on balanced datasets is also of interest and not explored. Finally, while RAC clearly demonstrates the benefit of an explicit retrieval component, the data being retrieved (labels) is of limited value and imposes a cap on RACs performance---a natural extension is to query for whole paragraphs or captions. However, the 76 token limit imposed by the CLIP text encoder prevents this, and would need to be increased. We leave this for future work.

\section{Conclusion}
We 
have 
introduced  RAC, a generic approach to augmenting standard classification pipelines with an explicit retrieval module. RAC's retrieval module, without prompting, achieves a high level of accuracy on tail classes, freeing up the base encoder to focus on common classes.  RAC improves upon the state-of-the-art results on the iNat  and Places365-LT benchmarks by a large margin for the task of long-tail image classification. 
We hope that RAC represents a step towards more effectively making use of external memory within common computer vision architectures, and we predict its use for other vision tasks, particularly, 
such as one/few shot learning, 
and 
continual learning without catastrophic forgetting. 

{\small
\bibliographystyle{ieee_fullname}
\bibliography{library}
}

\clearpage
\appendix

\section*{Appendix}

\renewcommand{\thefigure}{S\arabic{figure}}
\setcounter{figure}{0}
\renewcommand{\thetable}{S\arabic{table}}
\setcounter{table}{0}

We provide more information here.

\section{Additional Experiments}

\subsection{Re-weighting and Temperature Scaling}

\begin{table}[b!]
\centering
\begin{tabular}{@{}lrrrr@{}}
\toprule
\multicolumn{5}{c}{\textbf{Places365-LT}}                                                \\ \midrule
\textbf{Re-weighting} & \textbf{Many}  & \textbf{Med}   & \textbf{Few}   & \textbf{All}   \\ \midrule
None                 & \textbf{49.72} & 49.34          & 40.46          & 47.75          \\
Inverse sqrt.            & 47.12          & \textbf{49.58} & \textbf{47.65} & \textbf{48.32} \\
Inverse log              & 44.88          & 51.35          & 47.54          & 48.29          \\\midrule
\multicolumn{5}{c}{\textbf{iNat}}                                                        \\ \midrule
None                 & \textbf{75.92} & \textbf{80.48} & 81.07          & \textbf{80.24} \\
Inverse sqrt.            & 71.13          & 80.02          & 81.19          & 79.56          \\
Inverse log              & 67.17          & 80.04          & \textbf{81.67} & 79.35          \\ \bottomrule
\end{tabular}
\caption{Effect of common re-weighting schemes when combined with logit adjustment for RAC's top-level loss.}
\label{tbl:reweighting}
\end{table}

As a top level loss for RAC, we make use of unscaled logit adjustment exclusively, with no reweighting (i.e. $\alpha_y = 1 \quad \forall y\in\mathcal{Y}$) and no temperature scaling ($\tau =1$) in \cref{eq:general}. This loss is theoretically well-grounded \cite{lasm}, and is appealing due to its simplicity. Nevertheless, other works have noted that the combination of logit adjustment with re-weighting often leads to higher empirical performance \cite{LDAM}. While optimizing the top-level loss is not the focus of RAC, we include here a comparison of RAC performance under various re-weighting schemes when combined with logit adjustment for completeness. Specifically, we consider inverse log
\begin{equation}
    \alpha_y = \frac{1}{\log{N_y}}
\end{equation}
and inverse square-root 
\begin{equation}
    \alpha_y = \frac{1}{\sqrt{N_y}}
\end{equation}
class-frequency based re-weighting of individual sample losses. 

We confirm 
that 
the same effect is present for RAC on Places365-LT, with overall accuracy increasing with the use of both re-weighting schemes, with improvement most pronounced for tail classes (\cref{tbl:reweighting}). However, this trend does not hold on iNat. 

\begin{figure}
    \centering
    \includegraphics[width=0.9\columnwidth]{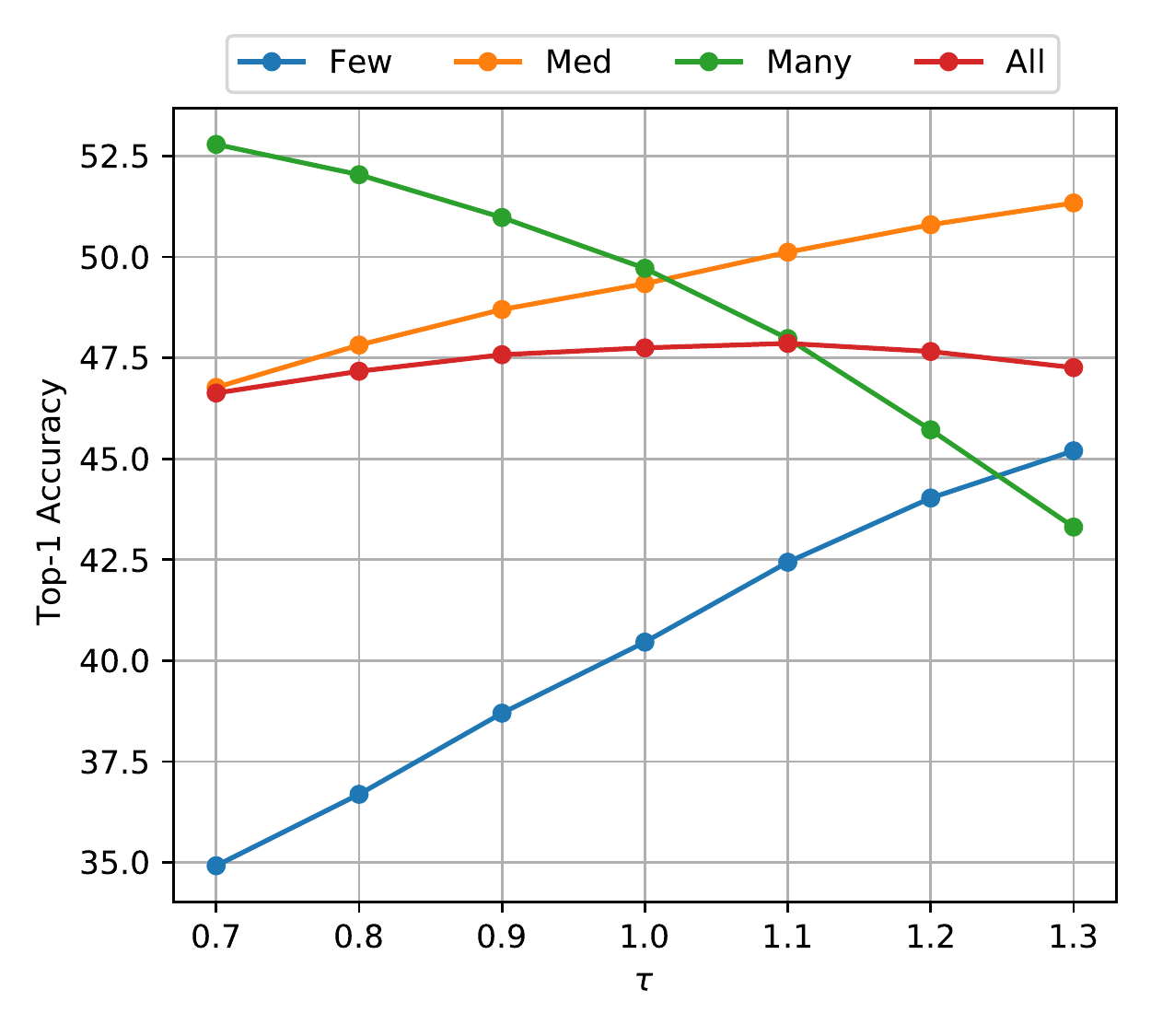}
    \caption{Effect of $\tau$ within the LACE loss on balanced performance on the Places365-LT dataset.}
    \label{fig:tausweep}
\end{figure}

For Places365-LT, we also we perform a sweep across $\tau$, to evaluate if the same effect can be achieved with manual temperature scaling (\cref{fig:tausweep}). Higher $\tau$ does result in slightly higher overall accuracy, however this effect is minor in comparison to re-weighting. This divergence from theory is likely due to the non-separability of many classes in Places365-LT due to the high label noise.

\subsection{Index Ablations}
\begin{table*}
\centering
\begin{tabular}{@{}rllrrrrrrr@{}}
\toprule
\multirow{2}{*}{\textbf{Index Size}} &
  \multicolumn{1}{c}{\multirow{2}{*}{\textbf{Index Type}}} &
  \multicolumn{1}{c}{\multirow{2}{*}{\textbf{Distance}}} &
  \multicolumn{5}{c}{\textbf{Test}} &
  \textbf{Train} &
  \multirow{2}{*}{\textbf{QT (ms/sample)}} \\
 &
  \multicolumn{1}{c}{} &
  \multicolumn{1}{c}{} &
  \textbf{Many} &
  \textbf{Med} &
  \textbf{Few} &
  \textbf{Top-1} &
  \textbf{Top-5} &
  \textbf{Top-1} &
   \\ \midrule
62.5k & Exact & $L_2$     & 39.97 & 26.74 & 18.65 & 29.91 & 53.81 & 99.97 & 1.12 \\
62.5k & Exact & Cosine & 39.84 & 26.51 & 18.03 & 29.64 & 53.31 & 99.96 & 1.14 \\
62.5k & HNSW  & $L_2$     & 39.89 & 26.51 & 18.03 & 29.66 & 53.32 & 99.79 & 1.06 \\
62.5k & HNSW  & Cosine & 39.57 & 26.26 & 17.68 & 29.37 & 52.83 & 99.79 & 1.07 \\
11.2M & Exact & $L_2$     & -     & -     & -     & -     & -     & -     & 7.96 \\
11.2M & HNSW  & $L_2$     & -     & -     & -     & -     & -     & -     & 3.01 \\ \bottomrule
\end{tabular}
\caption{Index ablations on Places365-LT. QT indicates Query Time. Large-sample indices are filled with the ImageNet21k dataset. }
\label{tbl:retab}
\end{table*}

We examine the effect of the distance metric and index type on RAC's lookup performance and speed in \cref{tbl:retab}. To quantify error induced by an approximate index, we include the lookup accuracy on the index content itself (training set) in addition to the validation accuracy. Query Time (QT) includes encoding samples with a ViT-B-16 model, which is the primarily overhead on small indexes.

We observe that the choice of distance metric ($\ell_2$ vs.\  cosine) has little effect, as may be expected, give the high dimentionality of the index. Cosine distance does introduce a minor computational overhead due to the need to normalize the embeddings prior to querying. The drop in accuracy due to use of an approximate (HNSW) instead of exact $k$-NN is also minor, but comes with a significant (2$\times$) speedup on large index's. Construction time is constant across all indexes at $0.02\mu$s per sample, with the exception of large-index HNSW, which requires $0.05\mu$ per sample. In summary, performance differences are minor when querying small indexes, however as the index size grows, the choice of HNSW becomes critical to ensure lookup time does not bottleneck training.

\subsection{Per-class Accuracy on iNat}
\begin{figure}[b]
\centering
\includegraphics[width=\columnwidth]{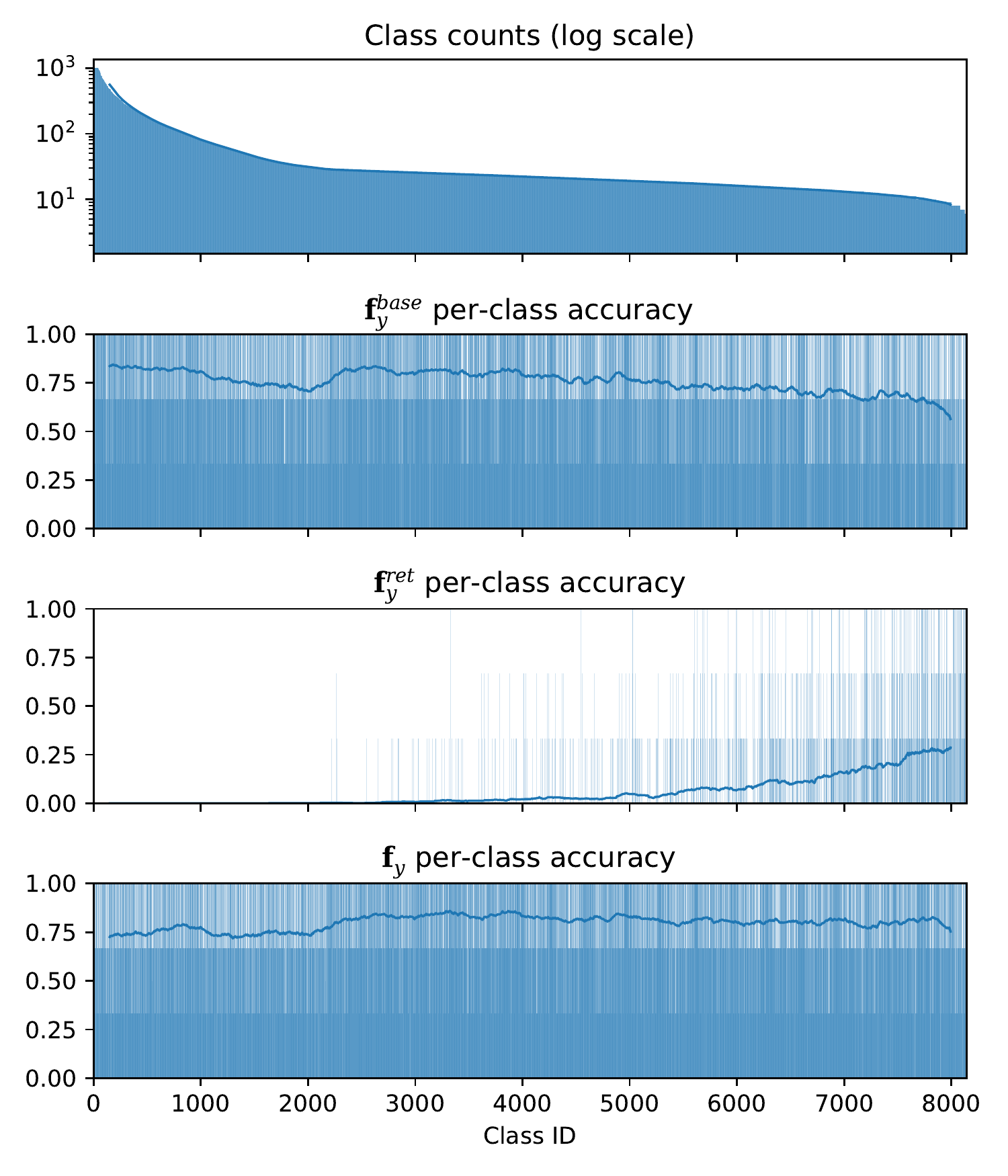}
\caption{Per-class top-1 accuracy on iNat from each branch's output. The $300$ sample moving average over classes (solid line) is shown for clarity.}
\label{fig:inatperclass}
\end{figure}

We include the same per-class visualization presented in the main body for Places365-LT (\cref{fig:Per-class}), for iNat. Note that for iNat only 3 samples are present for each class in the validation set, hence the square-wave appearance of the plots (\cref{fig:inatperclass}). Nevertheless, the same trend is clearly visible in the sliding window moving average, with the retrieval module performing best on tail classes and the base network largely focusing on the many and mid-frequency classes.

\section{Further Details}
For \textbf{Places365-LT} we use the training and validation splits during development and report final numbers on the test set, with no validation samples used during final training. For \textbf{iNaturalist2018}, following prior work, we report results on the released validation split, as labels for the test set are not publicly available. 

In the main text, we use several common model variants. While the architectures for these models are standard, we specify the high-level design choices in \cref{tbl:resnets,tbl:vits}. These choices are consistent across both the $224\times 224$ and $384 \times 384$ variants.

\begin{table}[]\label{tbl:dataset}
\centering
\begin{tabular}{@{}lrr@{}}
\toprule
                 & \textbf{Places365-LT} & \textbf{iNat} \\ \midrule
Samples (train)  & 62,500                & 437,513       \\
Samples (test)   & 36,500                & 24,426        \\
Classes          & 365                   & 8,142         \\
Imbalance Factor & 500                   & 996           \\ \bottomrule
\end{tabular}
\caption{Dataset Details}
\end{table}

\begin{table}[]
\centering
\begin{tabular}{@{}lrr@{}}
\toprule
\textbf{Hyperparameter} & \multicolumn{1}{c}{\textbf{ViT-B-16}} & \multicolumn{1}{c}{\textbf{ViT-B-32}} \\ \midrule
Patch size     & 16        & 32        \\
Depth          & 12        & 12        \\
Embedding Dimension & 768       & 768       \\
Attention Heads    & 12        & 12        \\
Parameters      & 85.8M     & 87.4M     \\ \bottomrule
\end{tabular}
\caption{ViT model architecture details.}
\label{tbl:vits}
\end{table}

\begin{table}
	\centering
	\begin{tabular}{@{}lrr@{}}
		\toprule
		\textbf{Hyperparameter}  & \textbf{RN50}    & \textbf{RN152d}     \\ \midrule
		Input size               & $244 \times 224 \times 3$         & $256 \times 256 \times 3$          \\
		Head                     & Avg. Pool, FC     & Avg. Pool, FC       \\
		Convolutions             & standard          & standard           \\
		Stem Convolutions        & 1 layer, $3\times 3$      & 3 layer, $3\times 3$        \\
		Stem width               & 32                & (128, 128, 128)       \\
		Layers                   & {[}3, 4, 6, 3{]}  & {[}3, 8, 36, 3{]}   \\
		Pool size               & $7\times 7$               & $8\times 8$ \\
		Num. features      &    2048               & 2048 \\
		Parameters               & 23.5M             & 58.2M             \\ \bottomrule
	\end{tabular}
	\caption{ResNet model architecture details. }
\label{tbl:resnets}
\end{table}

All training is carried out on $8\times$ $32$GB A$100$ GPU's. We largely follow the procedures outlined in \cite{howtotrainvit}, with the following alterations. We finetune with AdamW \cite{adamw} instead of SGD, and make use of low-magnitude RandAugment \cite{randaugment} alone for data augmentation, with no Color-Jitter, Mixup, Cutmix, RandomErase or Augmix applied. Full hyperparameters are shown in \cref{tbl:vithyperparams}. While we found the use of Mixup and Cutmix does boost performance on standard ViTs trained under a BalCE loss, their use of combined targets requires special treatment to make compatible with the LACE loss, which requires hard targets in order to assign the class adjustment. While one approach may be to apply the `merged' class adjustment, the performance benefit is marginal and hence we simply did not include either approach in RAC's data augmentation pipeline.

\begin{table}\label{tbl:trainingdataset}
\centering
\begin{tabular}{@{}lrr@{}}
\toprule
                 & \textbf{Places365-LT} & \textbf{iNat} \\ \midrule
Batch Size       & 200                   & 50            \\
Learning Rate    & 5e-5              & 1e-4      \\
Epochs           & 30                    & 20            \\ \bottomrule
\end{tabular}
\caption{Dataset specific training hyperparameters}
\end{table}

\begin{table}
	\centering
	\begin{tabular}{@{}lr@{}}
		\toprule
		\textbf{Hyperparameter}     & \textbf{Value}  \\ \midrule
		Global normalization means         & [0.5, 0.5, 0.5] \\
    	Global normalization stds          & [0.5, 0.5, 0.5] \\
    	Crop (train and test)       & 0.95    \\
		Distributed                 & DDP 8 GPU's    \\
		LR schedule                    & cosine         \\
		Min LR                        & 1e-7           \\
		Warmup LR                   & 1e-7              \\
		Warmup epochs               & 5              \\
		Optimizer                   & adamw           \\
		Beta 1                      & 0.9            \\
		Beta 2                      & 0.999          \\
		Eps.                         & 1e-8           \\
		Gradient clipping           & $L_2$ Norm           \\
		Gradient clipping magnitude & 1.0            \\
		RandAugment magnitude       & 1              \\
		RandAugment layers          & 3              \\
		RandAugment noise std.      & 0.5            \\
		Weight decay                & 0.02           \\
		Label smoothing             & 0.1            \\
		Stochastic depth            & 0.1            \\
		Random erase prob.          & 0.0           \\
		Color jitter                & 0.0            \\
		Random scale                & {[}0.75, 1.33{]} \\
		Random crop                 & No             \\
		Horizontal flip prob.       & 0.5   \\
		Random rotation             & No             \\
		Mixed precision level       & O2             \\ \bottomrule
	\end{tabular}
\caption{Training hyperparameters for the ViT models}
\label{tbl:vithyperparams}
\end{table}


\section{Retrieval Branch Visualization}\label{apx:vis}
\begin{figure*}    
    \centering
    \includegraphics[width=\textwidth]{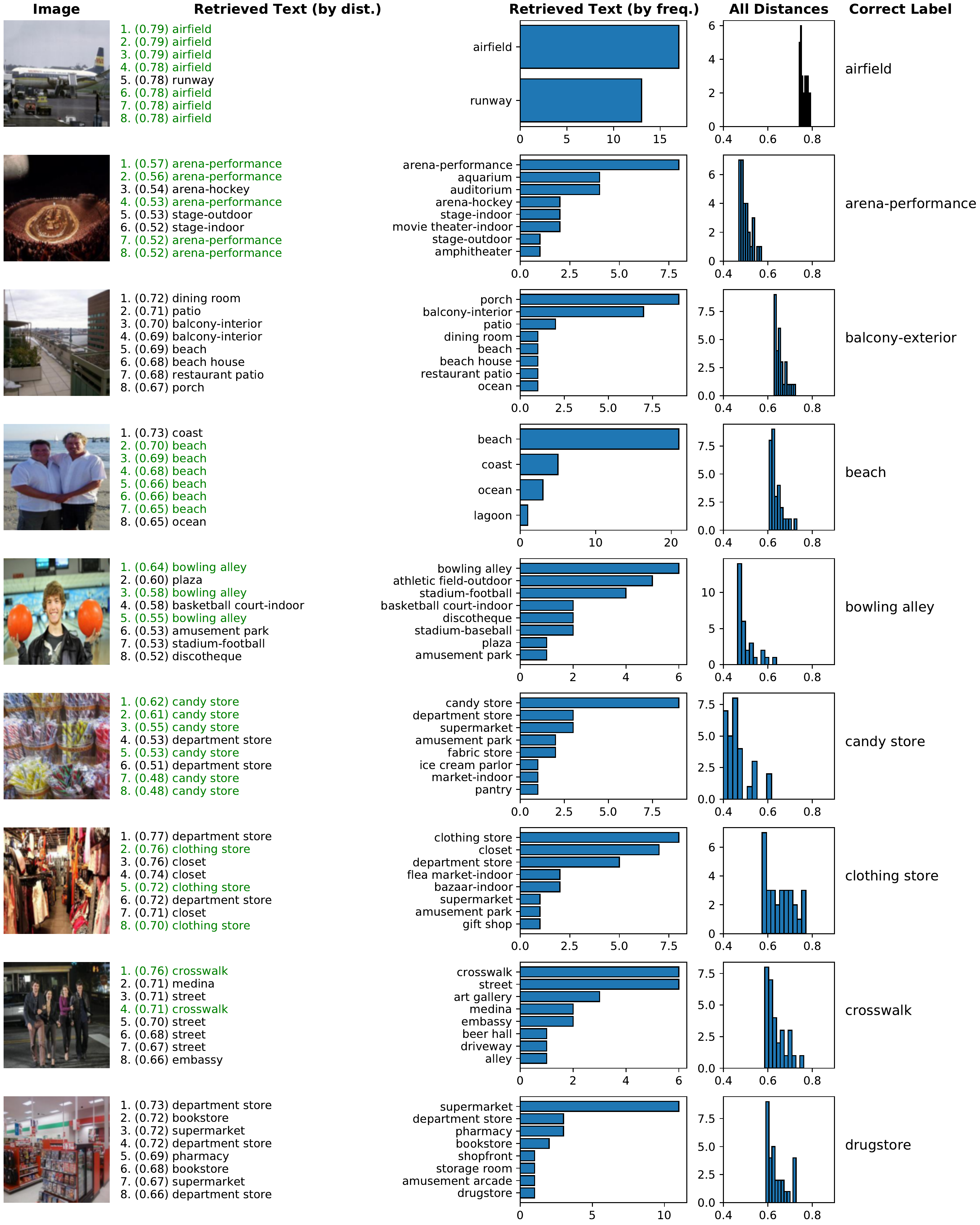}
    \caption{Retrieval branch visualization for randomly drawn samples from Places365-LT. For detailed explanation see \cref{apx:vis}.}
    \label{fig:placesExamples}
\end{figure*}

\begin{figure*}   
    \centering
    \includegraphics[width=\textwidth]{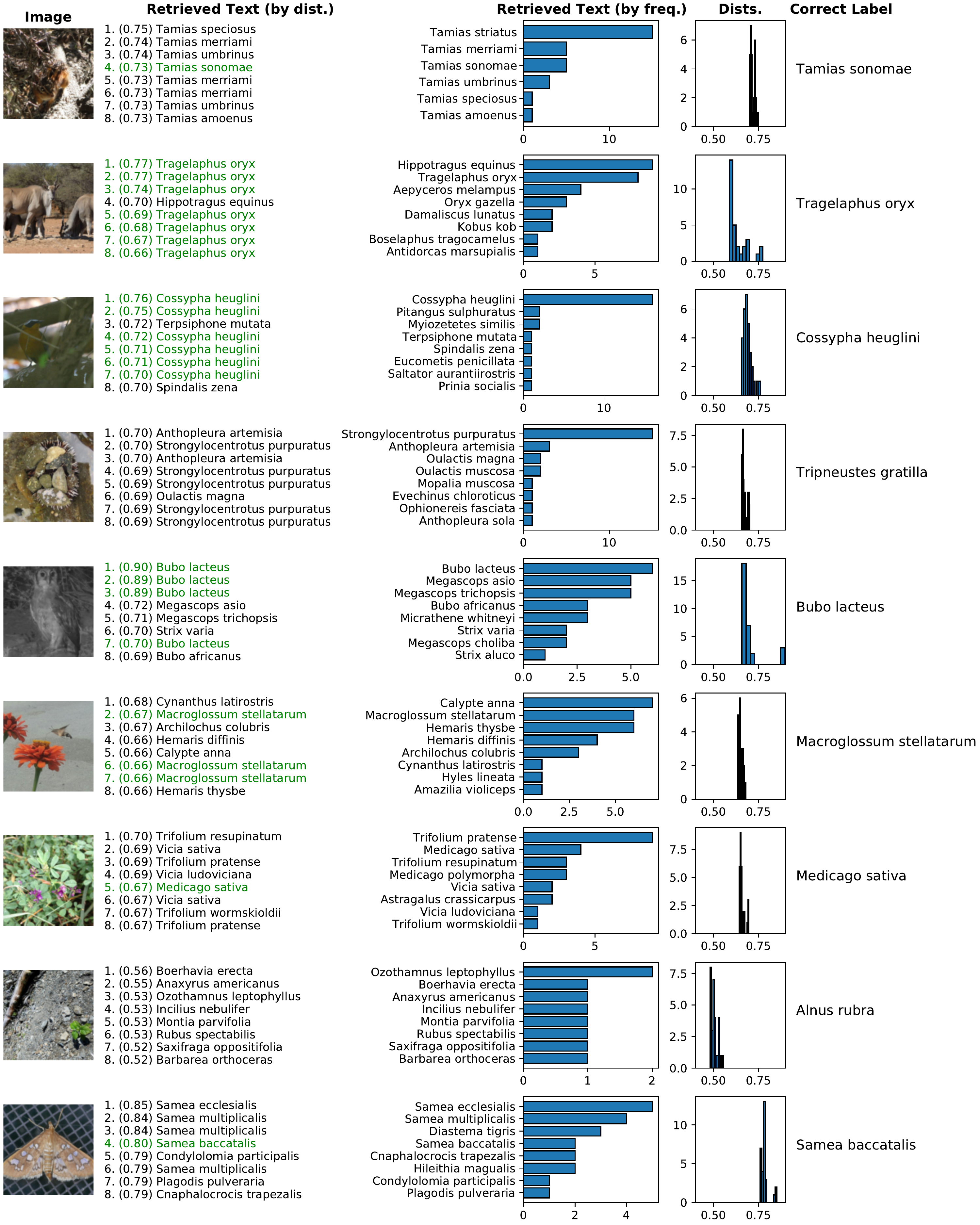}
    \caption{Retrieval branch visualization for randomly drawn samples from iNaturalist2018. For detailed explanation see \cref{apx:vis}.}
     \label{fig:iNatExamples}
\end{figure*}

While \cref{tbl:retrieval} quantifies top-1 performance of the retrieval branch, non-exact match snippets will also effect RAC as they are still likely to be informative. If `plane', `runway', `concrete', `sky', `propeller' etc. are returned for example, it is not difficult for $\mathbf{B}$ to place a high score on `airport'. We visualize the returned strings for random samples by distance and frequency in \cref{fig:placesExamples,fig:iNatExamples}. For all runs, $k=30$ as in the main work. 

Each column displays (from left to right):
\begin{enumerate}
\itemsep -0.0861cm
    \item 
 query image $x_q$,
 \item retrieved labels sorted by distance (distance shown in brackets, exact matches colored green),
 \item 
 retrieved labels and occurrence counts, sorted from most to least frequent,
 \item  histogram of distances to all returned samples,
 \item  the correct label. 

\end{enumerate}

We restrict the lists of retrieved labels to the top eight for visualization purposes. Note that as the distance metric is cosine, a higher value corresponds to more similar samples. 

\end{document}